\pdfoutput=1

\documentclass[11pt]{article}

\usepackage[preprint]{acl}

\usepackage{times}
\usepackage{latexsym}

\usepackage[T1]{fontenc}

\usepackage[utf8]{inputenc}

\usepackage{microtype}

\usepackage{inconsolata}

\usepackage{graphicx}

\usepackage{amsmath,amssymb,amsfonts}
\usepackage[linesnumbered,ruled,vlined]{algorithm2e}

\usepackage{textcomp}
\usepackage{xcolor}
\usepackage{enumitem}

\SetKwInput{KwInput}{Input}
\SetKwInput{KwOutput}{Output}
\usepackage{booktabs}
\usepackage{caption}
\usepackage{subcaption}
\usepackage{multirow}

\title{MDIT: A Model-free Data Interpolation Method \\ for Diverse Instruction Tuning}

\author{Yangning Li\footnotemark[2]~~~~Zihua Lan\footnotemark[2]~~~~Zihua Lan~~~~Lv Qingsong\\ \bf ~~~~Yinghui Li~~~~Hai-Tao Zheng\footnotemark[3]\\
   Tsinghua University}

\begin{document}
\maketitle
\begin{abstract}
As Large Language Models (LLMs) are increasingly applied across various tasks, instruction tuning has emerged as a critical method for enhancing model performance. However, current data management strategies face substantial challenges in generating diverse and comprehensive data, restricting further improvements in model performance. To address this gap, we propose MDIT, a novel model-free data interpolation method for diverse instruction tuning, which generates varied and high-quality instruction data by performing task interpolation. Moreover, it contains diversity-based clustering strategies to ensure the diversity of the training data. Extensive experiments\footnote{The code will be open source upon acceptance.} show that our method achieves superior performance in multiple benchmark tasks. The LLMs finetuned with MDIT show significant improvements in numerous tasks such as general question answering, math reasoning, and code generation. MDIT offers an efficient and automatic data synthetic method, generating diverse instruction data without depending on external resources while expanding the application potential of LLMs in complex environments.
\end{abstract}

\section{Introduction}
Instruction tuning has enabled large language models (LLMs) to accurately follow human instructions and significantly enhance their performance \cite{longpre2023flan, zhang2023instruction,yi2024survey}. The diversity of instruction datasets plays an essential role in improving LLM's ability to handle various scenarios \cite{muscato-etal-2024-overview, fan2025combatting}. Therefore, recent research focuses on curating high-diversity and wide-ranging instruction datasets \cite{mukherjee2023orca,chung2024scaling}.

In recent years, numerous studies attempt to increase the diversity of instruction datasets by filtering out simpler and less varied data \cite{liu2024what, pan2024g, tan-etal-2024-uva}. However, data selection methods primarily focus on removing low-diversity data and addressing the negative effects of overly simplistic data, but fail to expand the diversity of the original dataset fundamentally.

To overcome the limitations of data selection methods, researchers turn to data synthesis \cite{xu2024wizardlm, zhao2024tree, chen2024dog}, generating diverse instruction data to improve the capacity of LLM for handling complex tasks. For example, Self-Instruct \cite{wang2022self} uses some human-annotated examples to prompt the model into creating more varied datasets, while UltraChat \cite{ding2023enhancing} iteratively refines multi-turn dialogues through systematically designed prompts. However, data synthesis heavily depends on external models and extensive human annotation, leading to high labor costs and inconsistent annotation quality.

It leads to a critical question naturally: how to effectively expand the diversity of instruction data and enhance the performance of LLM without relying on additional external models?

To address this challenge, Mixup \cite{zhang2018mixupempiricalriskminimization}, originally proposed in the computer vision domain, provides a promising approach by linearly blending images and their corresponding labels to improve model robustness and generalization. However, directly applying Mixup to instruction tuning in LLMs is tough due to the fundamental differences between structured image-label pairs and complex, natural language instructions. Traditional Mixup methods primarily perform simple linear interpolations within the same task, which does not naturally extend to the diverse, multi-faceted nature of instruction datasets.

To this end, we propose \textbf{MDIT}, a \textbf{M}odel-free \textbf{D}ata \textbf{I}nterpolation method for \textbf{D}iverse \textbf{I}nstruction \textbf{T}uning. 
Concretely, we (1) apply interpolation on different tasks at the embedding layer to generate more diverse tasks and (2) use clustering to filter out low-diversity data. To achieve this, we first transform samples into hidden states within the model, then perform linear interpolation on the embeddings to create new tasks, thereby fundamentally enhancing data diversity. Next, a clustering step ensures the overall diversity of the training data without relying on additional resources.

The key innovation of our MDIT over existing data synthesis methods is its labor-free nature, as it avoids the need for external resources to minimize costs. By avoiding reliance on pretrained models or manual annotations, our method reduces potential errors and ensures robust and diverse data fusion.

We conduct comprehensive experiments on several benchmarks including ARC Challenge, MMLU-Math, Humaneval, and MBPP, showing that our MDIT significantly enhances LLM performance. Furthermore, it outperforms SOTA data selection and synthesis methods by generating more diverse tasks while discarding external resources.

The key contributions of this paper as follows:
\begin{itemize}
    \item We analyze current instruction data management strategies systematically, revealing that data selection methods fail to expand diversity basically, while data synthesis methods often rely on additional resources.
    \item We propose MDIT, a model-free data interpolation method that generates diverse tasks without external resources, improving the overall performance of LLM.
    \item Extensive experiments across multiple benchmarks show the effectiveness of MDIT, achieving superior results without the need for additional resources.
\end{itemize}

\section{Related Work}

\subsection{Instruction Data Management for Diversity}
Recent research on managing instruction data diversity can be classified into filter-based data selection and generation-based data synthesis methods.

\subsubsection{Instruction Data Selection}
Data selection methods aim to filter out and remove low-diversity instruction data, including metric-based and model-based methods~\cite{chen2023star,qiu2024ease}. Metric-based selections \cite{gonen2022demystifying, zhou2023dataset,zeng-etal-2025-data} use quantitative metrics to identify diverse instruction data. Instruction mining \cite{cao2023instruction} uses a linear equation to evaluate instruction quality, while InstructionGPT-4 \cite{wei2023instructiongpt} further filters multimodal instruction data~\cite{yu2024recent,chen2024deep} according to CLIP scores \cite{radford2021learning} and instruction length. Model-based selections \cite{wu2023self,chen2023alpagasus,yu2023wavecoder, ge-etal-2024-clustering} leverage LLMs as data selectors to identify more diverse instructions \cite{li-etal-2024-quantity, liu2024what}. INSTAG \cite{lu2023instag} utilizes ChatGPT to annotate instruction data. Active Instruction Tuning \cite{kung2023active} filters tasks based on prompt uncertainty, while Nuggets \cite{li-etal-2024-one} employs two-stage scoring to select diverse data. However, these data selection methods focus on filtering out low-diversity data and fall short of fundamentally enriching the instruction dataset by adding novel instructions.

\subsubsection{Instruction Data Synthesis}
Data synthesis methods aim to generate diverse instruction data and improve the robustness of LLMs. Some work leverages the generative capabilities of LLMs to create new instructions \cite{taori2023stanford, he-etal-2024-community, kou-etal-2024-knn}, utilizing semantic parsing \cite{zhao2024tree}, transforming simple queries into complex tasks \cite{xu2024wizardlm} and blending model outputs with human-written content \cite{chen2024dog}, effectively enhancing dataset diversity and quality. However, these data synthesis methods typically depend on powerful external models or extensive human annotation, leading to high computational costs and potential data leakage risks. Different from them, our MDIT is entirely labor-free, generating diverse tasks without external resources. By incorporating diversity-based clustering, we further ensure the variety of the training data.

\subsection{Mixup Methods in Computer Vision}
To enhance data diversity, \cite{zhang2018mixupempiricalriskminimization} introduced Mixup for computer vision, which creates new training samples by linearly interpolating pairs of input images and their corresponding labels. Numerous Mixup variants \cite{yun2019cutmix, qin2020resizemix, kim2020puzzle, chen2022stackmix, wang2024enhance, sun2024patch, islam2024diffusemix} show improvements in tasks such as image classification and object detection, highlighting their effectiveness in enhancing data diversity and model robustness. However, these Mixup methods primarily focus on blending samples from similar categories within the same task. Significantly different from them, our MDIT performs interpolation across multiple tasks, generating more diverse and dynamic training data and enhancing LLM's ability to handle complex challenges.

\begin{figure*}[ht]
    \centering
    \includegraphics[width=\textwidth,keepaspectratio]{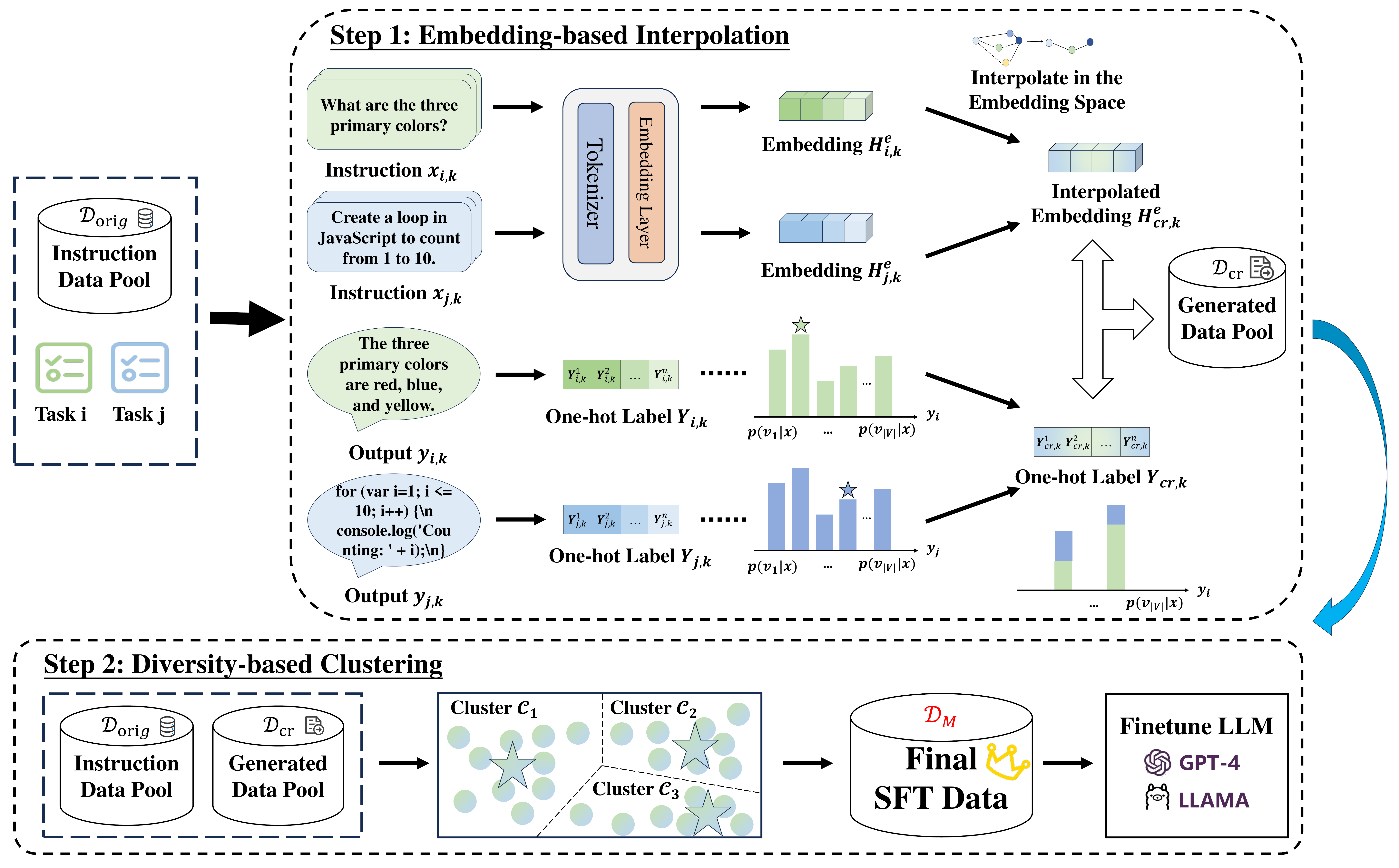}
    \caption{The framework of our method MDIT. MDIT consists of two primary steps: Embedding-based Interpolation and Diversity-based Clustering: In the first step, we perform task interpolation within the high-dimensional embedding space, generating new tasks that capture diverse semantic relationships. The second step involves clustering filtering to the curated set and selecting diverse training data from each cluster for instruction tuning.}
    \label{fig}
\end{figure*}

\section{Methodology}

We present the framework of \textbf{MDIT} in Figure \ref{fig}. Our method aims to select diverse training data for instruction tuning, consisting of two core phases: embedding-based task synthesis with interpolation (\S~\ref{sec:sec3.2}) and diversity-based data selection with clustering (\S~\ref{sec:sec3.3}). For the initial phase, we apply embedding interpolation across different tasks to create varied training tasks. Then we combine original and generated tasks and utilize clustering to ensure training data diversity. Finally, selected embeddings are directly used for LLM training, improving LLM's performance and robustness.

\begin{algorithm*}[htbp]
\caption{Embedding-Based Interpolation}
\label{alg:algorithm}
\KwInput{Training tasks \(\mathcal{D}_i \) and \(\mathcal{D}_j \), Interpolation weight $\alpha$}
\KwOutput{Augmented task $\mathcal{D}_{\text{cr}}$}
Initialize an empty dataset $\mathcal{D}_{\text{cr}}$\;
\ForEach{$x \in \mathcal{D}_i, \mathcal{D}_j$}{
    Tokenize the instruction $x$ into tokens: $\mathcal{T}_x$\;
    Embed the tokens into high-dimensional embeddings: $\mathbf{H}^{e}$\;
    One-Hot Encode the label $y \in \mathcal{D}_i, \mathcal{D}_j$: $\mathbf{Y}$\;
    Group samples based on similar input lengths\;
    \ForEach{pair of samples  \((\mathbf{H}_{i,k}^{e}, \mathbf{Y}_{i,k})\) , \((\mathbf{H}_{j,k}^{e}, \mathbf{Y}_{j,k})\) with similar input lengths}{
        Sample $\lambda$ from beta distribution with parameter $\alpha$: $\lambda \leftarrow \text{Beta}(\alpha, \alpha)$\;
        Generate Interpolated Embedding: $\mathbf{H}_{cr,k}^{e} = \lambda \cdot \mathbf{H}_{i,k}^{e} + (1 - \lambda) \cdot \mathbf{H}_{j,k}^{e}$\;
        Generate Interpolated Label: $\mathbf{Y}_{cr,k}^{n} = \lambda \cdot \mathbf{Y}_{i,k}^{n} + (1 - \lambda) \cdot \mathbf{Y}_{j,k}^{n}$\;
    }
    Add the generated embedding $\mathbf{H}_{cr,k}^{e}$ and the generated label $\mathbf{Y}_{cr,k}$ to the augmented task $\mathcal{D}_{\text{cr}}$\;
}
\Return{$\mathcal{D}_{\text{cr}}$}\;
\end{algorithm*}

\subsection{Preliminaries}
Mixup \cite{zhang2018mixupempiricalriskminimization} is a data augmentation technique originally developed for computer vision, designed to enhance model's generalization capabilities. The core idea of Mixup is to perform linear combinations on both input data and labels during the training process. By linearly combining two distinct training samples and their corresponding labels in specific proportions, new mixed samples are generated, which increases training data diversity. Mixup enables the model to encounter various intermediate states within the data space during training, rather than relying solely on the original data. Specifically, given input data samples \(x_i\) and \(x_j\) , along with their corresponding labels \(y_i\) and \(y_j\) , Mixup creates new training samples $x_{\text{new}}$ and labels $y_{\text{new}}$ by performing a linear combination of the two pairs \((x_i, y_i)\) and \((x_j, y_j)\)

\[
x_{\text{new}} = \lambda x_i + (1 - \lambda) x_j
\tag{1}
\]
\[
y_{\text{new}} = \lambda y_i + (1 - \lambda) y_j
\tag{2}
\]

\noindent where $\lambda$ is randomly sampled from the beta distribution. The generated samples are then used to train the neural network, which learns a more comprehensive range of data distribution characteristics through data augmentation, thereby improving its performance on unseen data.

\subsection{Embedding-Based Interpolation} \label{sec:sec3.2}

In this section, we detail our task-level interpolation method for generating diverse training data, aiming to expand the task distribution and improve the generalization ability of LLM, as summarized in Algorithm \ref{alg:algorithm}. MDIT performs interpolation between different tasks in a high-dimensional embedding space to create new tasks. These new tasks are generated by blending knowledge from multiple task distributions.

We define the training sets \(\mathcal{D}_i \) and \(\mathcal{D}_j \) for task $i$ and task $j$ as $\mathcal{D}_i = (\mathbf{X}_i, \mathbf{Y}_i) = \left\{ \left( x_{i,k}, y_{i,k} \right) \right\}_{k=1}^{N_i}$ and $\mathcal{D}_j = (\mathbf{X}_j, \mathbf{Y}_j) = \left\{ \left( x_{j,k}, y_{j,k} \right) \right\}_{k=1}^{N_j}$. Concretely, a LLM \(f_\theta\) consists of \(\mathcal{L}\) layers, and the hidden representation of samples \(x_{i,k}\) at the embedding layer is denoted as \(\mathbf{H_{i,k}^e} = f_{\theta}(x_{i,k})\). The samples from task $i$ and task $j$ are mapped into the high-dimensional embedding space through the model's embedding layer, while their corresponding labels are encoded as one-hot vectors, getting $\mathcal{D}_i^e = (\mathbf{H}_i^{e}, \mathbf{Y}_i)$ and $\mathcal{D}_j^e = (\mathbf{H}_j^{e}, \mathbf{Y}_j)$.

Next, we apply the task interpolation separately in the high-dimensional embedding space. First, an interpolation weight $\lambda \sim \text{Beta}(\alpha, \alpha)$ is randomly sampled from a Beta distribution with hyperparameter \(\alpha\) that controls the concentration of the distribution, the probability density function as follows: 

\[
f(\lambda; \alpha, \alpha) = \frac{\Gamma(2\alpha)}{\Gamma(\alpha)\Gamma(\alpha)} \lambda^{\alpha - 1} (1 - \lambda)^{\alpha - 1}
\tag{3}
\]

Then, we apply task interpolation to the hidden representations of two samples from different tasks and their corresponding labels \((\mathbf{H}_{i,k}^{e}, \mathbf{Y}_{i,k})\) and \((\mathbf{H}_{j,k}^{e}, \mathbf{Y}_{j,k})\) to generate new tasks as:
\begin{align}
\mathbf{H}_{cr,k}^{e} &= \lambda \cdot \mathbf{H}_{i,k}^{e} + (1 - \lambda) \cdot \mathbf{H}_{j,k}^{e} \tag{4} \\
\mathbf{Y}_{cr,k}^{n} &= \lambda \cdot \mathbf{Y}_{i,k}^{n} + (1 - \lambda) \cdot \mathbf{Y}_{j,k}^{n} \tag{5}
\end{align}
\noindent where the superscript "e" means "interpolation in the embedding space" while the subscript "cr" indicates "cross". \(\mathbf{H}_{i,k}^{e}\) represents the hidden representations of the $k$-th sample in the $i$-th task, while $\mathbf{Y}_{i,k}^{n}$ represents the label of the $k$-th sample in the $i$-th task, where each label length is $n$. The generated dataset retains the semantic information from the original dataset while incorporating randomness through the interpolation weights, enhancing the semantic diversity of the dataset. By applying the interpolation operation across multiple tasks, we generate a diverse set of tasks, which can be formally defined as:
\[
\mathcal{D}_{\text{cr}}^e = \left\{ \left( \mathbf{H}_{\text{cr,(k)}}^{e}, \mathbf{Y}_{\text{cr,(k)}} \right) \mid 1 \leq k \leq N_\text{cr}  \right\}
\tag{6}
\]
where \(N_\text{cr}\) is the number of samples in the generated tasks, and \(\mathbf{H}_{\text{cr,(k)}}^{e}\) represents the hidden representations of the generated samples while \(\mathbf{Y}_{\text{cr,(k)}}\) represents the corresponding labels.

Through task-level interpolation, we effectively expand the task distribution, introducing a wider variety of tasks into the training dataset, while improving the robustness and flexibility of LLMs.

\subsection{Diversity-Based Clustering} \label{sec:sec3.3}
After generating new tasks through embedding-based interpolation, it is essential to implement effective filtering strategies to eliminate low-diversity data from new tasks. Data selection ensures a diverse and high-quality training dataset, providing an optimal foundation for finetuning LLMs.

Concretely, we apply a clustering selection to ensure training dataset diversity. First, we combine the original dataset \( \mathcal{D}_{\text{orig}} \) and the generated dataset \( \mathcal{D}_{\text{cr}}^e \) to form a total dataset \( \mathcal{D}_{\text{total}}^e = \mathcal{D}_{\text{orig}}^e \cup \mathcal{D}_{\text{cr}}^e \). The combined dataset \( \mathcal{D}_{\text{total}}^e \) provides a comprehensive pool for clustering.

Then, the K-Means algorithm partitions \( \mathcal{D}_{\text{total}}^e \) into \( m \) clusters, optimizing the division by minimizing the sum of squared Euclidean distances $d$ between each data point and its corresponding cluster center \( c_m \). The set of clusters \( \mathcal{C} \) and the objective function $ f $ are defined as follows:

\[
\mathcal{C} = \textit{KMeans}(\mathcal{D}_{\text{total}}^{e}, m)
\tag{7}
\]
\[
f(\{c_m\}; \mathcal{C}) = \min_{\{c_m\}} \sum_{m=1}^{N_\mathcal{C}} \sum_{\mathbf{H}_{\text{cr, k}}^{e} \in \mathcal{C}_m} \| \mathbf{H}_{\text{cr, k}}^{e} - c_m \|_2^2
\tag{8}
\]
where \( c_m \) represents the center of the \( m \)-th cluster, \( \mathcal{C}_m \) is the set of data belonging to the \( m \)-th cluster, and \( \mathbf{H}_{\text{cr, k}}^{e} \) is the hidden representation of the \( k \)-th sample from the interpolation task. The dataset \( \mathcal{D}_{\text{total}}^{e} \) is divided into distinct clusters \( \mathcal{C} \) based on clustering results. After clustering, we compute the Euclidean distance $ d $ from each data point \( \mathbf{H}_{\text{cr, k}}^{e} \) to its respective cluster center \( c_m \), defined as \( d(\mathbf{H}_{\text{cr, k}}^{e}, c_m) = \| \mathbf{H}_{\text{cr, k}}^{e} - c_m \|_2 \). We select data points that \( d(\mathbf{H}_{\text{cr, k}}^{e}, c_m) \) is minimized, focusing on those closer to the cluster centers \( \mathcal{C} \) or within densely populated regions of the cluster. 

Finally, the total dataset \( \mathcal{D}_{\text{total}}^{e} \) is selected to a new dataset \( \mathcal{D}_{\text{M}}^{e} \), which enhances the coverage and informational value of the training data. The filtered dataset \( \mathcal{D}_{\text{M}}^{e} = (\mathbf{H}_{\text{M}}^{e}, \mathbf{Y}_{\text{M}})\) serves as the final training dataset for finetuning LLM, ensuring LLM learns from a diverse and representative set of tasks.

By combining embedding-based interpolation with diversity-based clustering, MDIT greatly expands training data diversity, thereby enhancing the generalization ability of LLMs. Additionally, MDIT provides an automatic data synthetic solution, enriching the diversity of instruction data without relying on external resources.

\begin{table*}[htb]
\scalebox{0.92}{
\centering
\begin{tabular}{lccccc}
\toprule

\textbf{Model} & \multicolumn{1}{c}{\textbf{General QA}} & \multicolumn{1}{c}{\textbf{Math Reasoning}} & \multicolumn{2}{c}{\textbf{Code Generation}} \\ 
& \textbf{ARC Challenge} & \textbf{MMLU-Math} & \textbf{HumanEval}  & \textbf{MBPP} & \textbf{Average}  \\
\midrule
\multicolumn{6}{c}{\textbf{Model FineTuned based on Sheared-LLaMa-1.3B}} \\
\midrule
Zero-Shot & 29.10 & 24.30 & 0.00 & 0.20 & 13.40 \\
IFD \cite{li-etal-2024-quantity} & 30.20 & 25.70 & 1.83 & 0.04 & 14.44 \\
DEITA \cite{liu2024what} & 29.61 & 23.60 & 2.44 & 0.04 & 13.92 \\
Evol-Instruct \cite{xu2024wizardlm} & 28.67 & 22.70 & 7.32 & 0.28 & 14.74 \\
\midrule
\textbf{MDIT (ours)} & 26.28 & 29.20 & 3.05 & 4.32 & \textbf{15.71} \\
w/o Cluster-Selection & 28.33 & 25.80 & 3.05 & 5.55 & 15.68 \\	
\midrule\midrule

\multicolumn{6}{c}{\textbf{Model FineTuned based on LLaMa-2-7B}} \\
\midrule
Zero-Shot & 44.11 & 30.50 & 16.46 & 17.68 & 27.19 \\
IFD \cite{li-etal-2024-quantity} & 47.10 & 30.20 & 21.95 & 18.59 & 29.46 \\
DEITA \cite{liu2024what} & 45.31 & 30.40 & 20.73 & 19.93 & 29.09 \\
Evol-Instruct \cite{xu2024wizardlm} & 43.60 & 24.60 & 25.61 & 19.05 & 28.22 \\		
\midrule
\textbf{MDIT (ours)} & 45.40 & 32.70 & 23.17 & 20.76 & \textbf{30.51} \\
w/o Cluster-Selection  & 46.25 & 31.80 & 22.56 & 20.84 & 30.36 \\
\midrule\midrule

\multicolumn{6}{c}{\textbf{Model FineTuned based on LLaMa-2-13B}} \\
\midrule
Zero-Shot & 50.17 & 33.90 & 22.56 & 16.63 & 30.81 \\
IFD \cite{li-etal-2024-quantity} & 49.57 & 35.00 & 26.22 & 25.36 & 34.04 \\
DEITA \cite{liu2024what} & 49.23 & 31.80 & 28.66 & 24.12 & 33.45 \\
Evol-Instruct \cite{xu2024wizardlm} & 49.57 & 34.00 & 28.66 & 25.00 & 34.31 \\		
\midrule
\textbf{MDIT (ours)} & 51.37 & 34.90 & 27.44 & 25.13 & \textbf{34.71} \\
w/o Cluster-Selection  & 50.43 & 32.30 & 28.05 & 26.63 & 34.35 \\

\bottomrule
\end{tabular}}
\caption{Evaluation Results on the Open LLM Leaderboard. We present the comparison results of our method MDIT with various baselines on \textsc{Sheared-LLaMa-1.3B}, \textsc{LLaMa-2-7b} and \textsc{LLaMa-2-13b}. We report the results of our MDIT and MDIT w/o cluster selection, the best overall performance in each group is in \textbf{bold}.}
\label{tab:main_experiment}
\end{table*}

\subsection{Model Training}
We use the selected dataset \( \mathcal{D}_{\text{M}} \) to finetune LLM. During training, LLM performs forward propagation to generate predictions, which are then compared to the true labels to compute the loss. The loss function $\mathcal{L}$ is defined as:

\[
\mathcal{L} = - \frac{1}{\mathcal{N}_{\text{M}}} \sum_{k,n,r} \log\left( P(\mathbf{Y}_{\text{M,k}}^{n}=r \mid \mathbf{H}_{\text{M,k}}^{e, n}) \right)  
\tag{9}
\]

\noindent where \( N_\text{M} \) is the total number of training data, including selected original and generated data. \( \mathbf{Y}_{\text{M,k}}^{n} \) is the true one-hot label and \( P(\mathbf{Y}_{\text{M,k}}^{n} = r \mid \mathbf{H}_{\text{M,k}}^{e, n}) \) is the predicted probability for the \(n\)-th token in the \( k \)-th sample.

Once the loss is computed, the model parameters are updated with the following gradient update rule: 
\[
\small
\theta \leftarrow \theta - \eta \cdot \frac{1}{\mathcal{N}_{\text{M}}} \sum_{k=1}^{\mathcal{N}_{\text{M}}} \left( P(\mathbf{Y}_{\text{M,k}}^{n}=r \mid \mathbf{H}_{\text{M,k}}^{e, n}) - \mathbf{Y}_{\text{M,k}}^{n} \right) \cdot \frac{\partial z_k^n}{\partial \theta}
\tag{10}
\]
where \( \theta \) represents the model parameters at the current iteration, \( \eta \) is the learning rate that controls the size of the parameter update, \( \frac{\partial z_k^n}{\partial \theta} \) is the gradient of the logit \( z_k^n \) relative to the model parameters \( \theta \).
The gradient update ensures the parameters are adjusted to minimize the loss, allowing LLM to improve its predictions with each iteration.

During finetuning, MDIT utilizes selected data to enhance training efficiency. Training on diverse tasks enables LLM to learn richer expressions, improving its performance on complex tasks.

\section{Experiments}

\subsection{Experiment Setup} 
\textbf{Datasets.} 
We use the general question-answering task Alpaca \cite{taori2023stanford}, the math reasoning task GSM8K \cite{cobbe2021training}, and the code generation task CodeAlpaca \cite{codealpaca} for training. To evaluate model performance, we adopt general question answering, math reasoning, and code generation benchmarks for automatic evaluation including ARC Challenge \cite{clark2018think}, MMLU-Math \cite{hendrycks2020measuring}, Humaneval \cite{chen2021evaluating}, and MBPP \cite{austin2021program}.

\noindent \textbf{Baselines.} We compare our MDIT method with several leading data selection and data synthesis methods. We consider the following baselines:

IFD \cite{li-etal-2024-quantity} selects a balanced subset of instructions by assessing the complexity of instructions through difficulty scores.

DEITA \cite{liu2024what} combines complexity and quality scoring models to evaluate the diversity and difficulty of each instruction, applying a nearest-neighbor distance threshold to maintain a varied and high-quality training set.

Evol-Instruct \cite{xu2024wizardlm} leverages the generative capabilities of LLMs to transform simple instructions into more complex variants.

For the baseline methods, we adopt the best parameters as reported in the original papers.

\noindent \textbf{Implementation Details}. We perform full-parameter finetuning on the \textsc{Sheared-LLaMA-1.3B} model \cite{xia2023sheared}, while using LoRA finetuning for \textsc{Llama-2 7B and 13B} \cite{touvron2023llama}. To ensure a fair comparison, we use the same setting for all finetuning experiments. The finetuning process lasts for 3 epochs with learning rate $\eta =  2\textit{e} - 5$ and a global batch size of 16. For MDIT, we set \( \alpha \) = 8 to sample \( \lambda \) from the Beta distribution, and set the number of generated samples per original sample pair $T = 1$.

\begin{figure*}[htbp]
\centering
\begin{subfigure}[b]{0.32\textwidth}
    \centering
    \includegraphics[width=\textwidth, height=4cm, keepaspectratio]{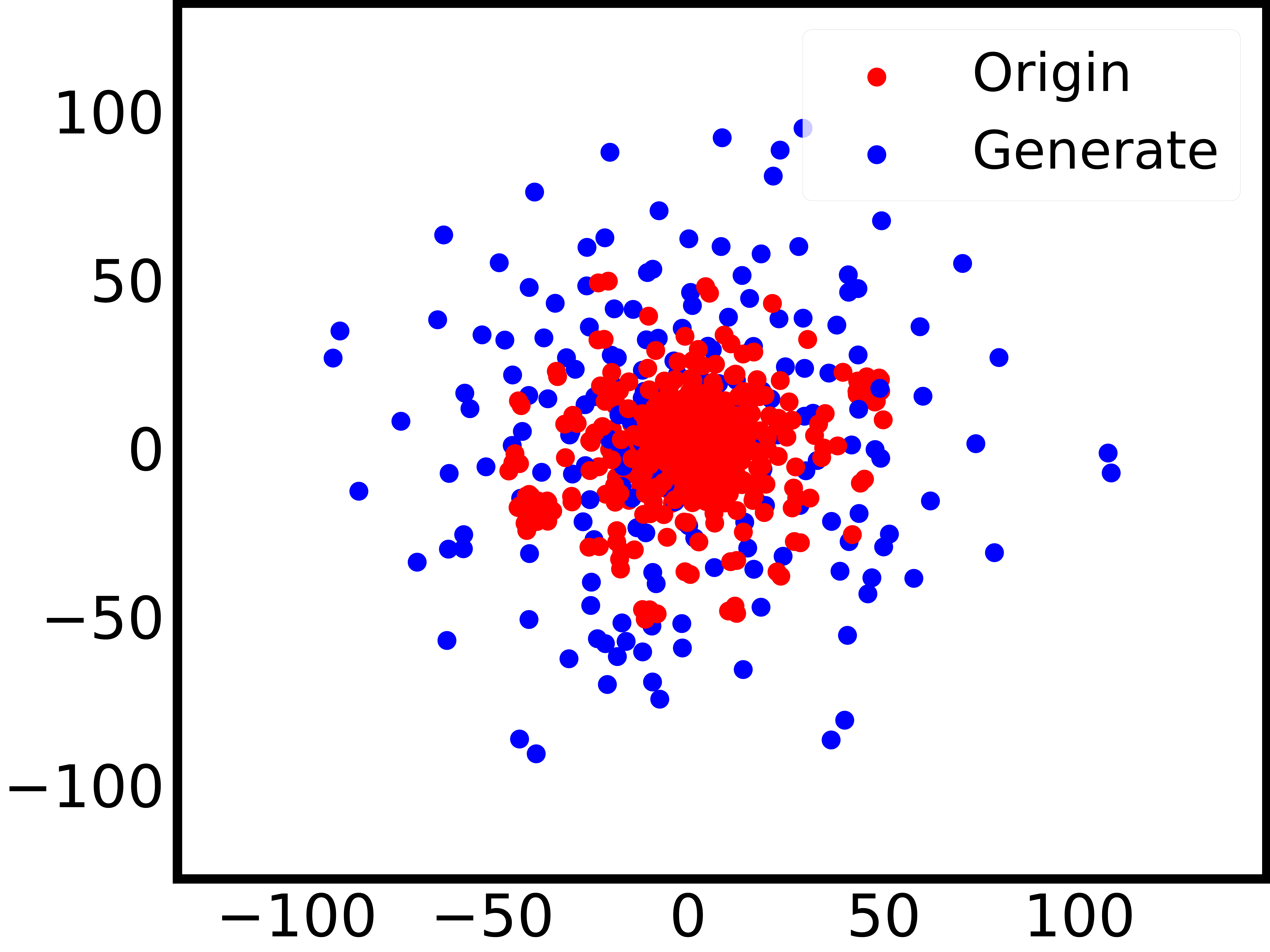}
    \caption{t-SNE visualizations}
\end{subfigure}
\begin{subfigure}[b]{0.32\textwidth}
    \centering
    \includegraphics[width=\textwidth, height=4cm, keepaspectratio]{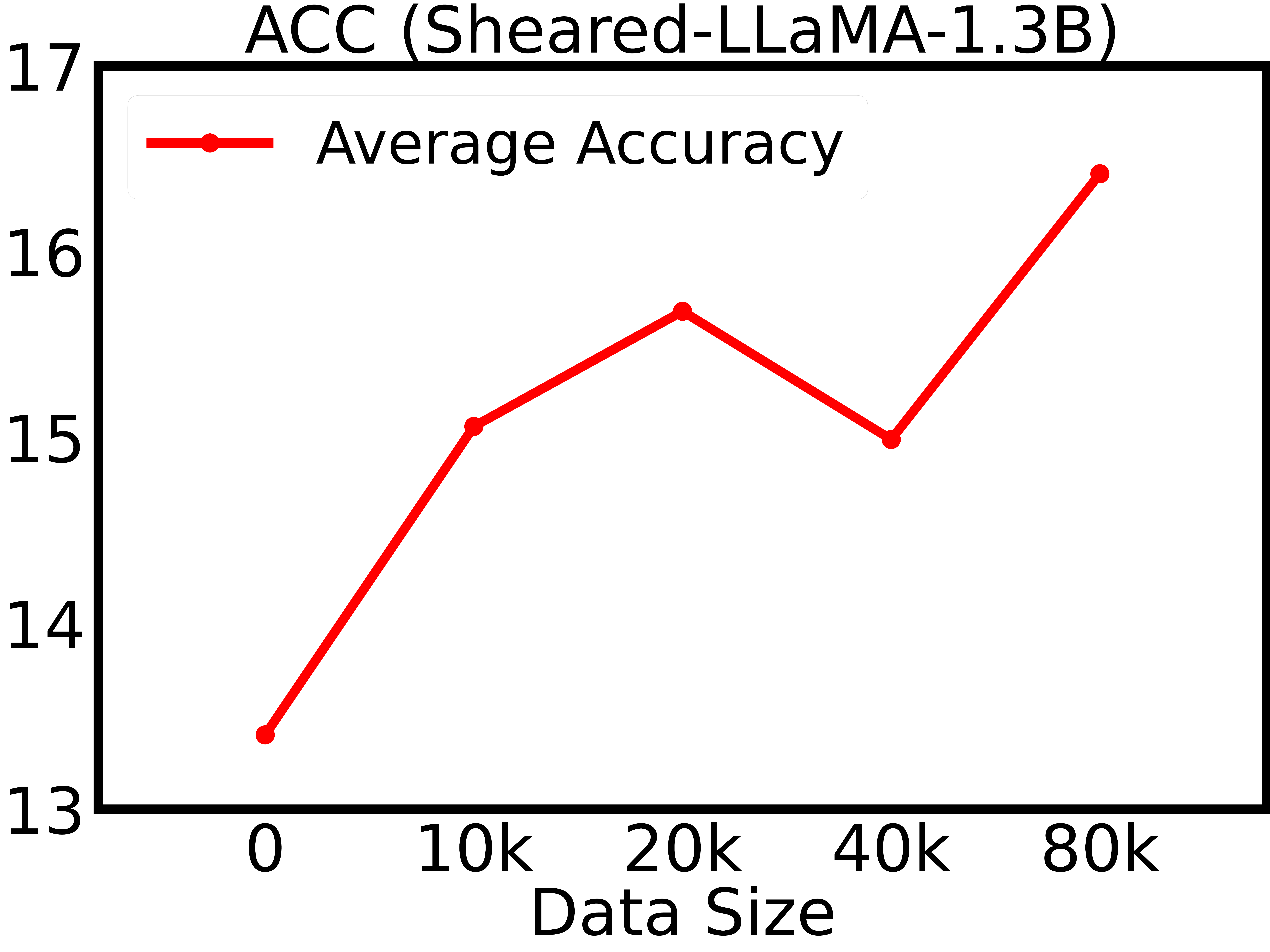}
    \caption{MDIT Scaling(1.3b)}
\end{subfigure}
\begin{subfigure}[b]{0.32\textwidth}
    \centering
    \includegraphics[width=\textwidth, height=4cm, keepaspectratio]{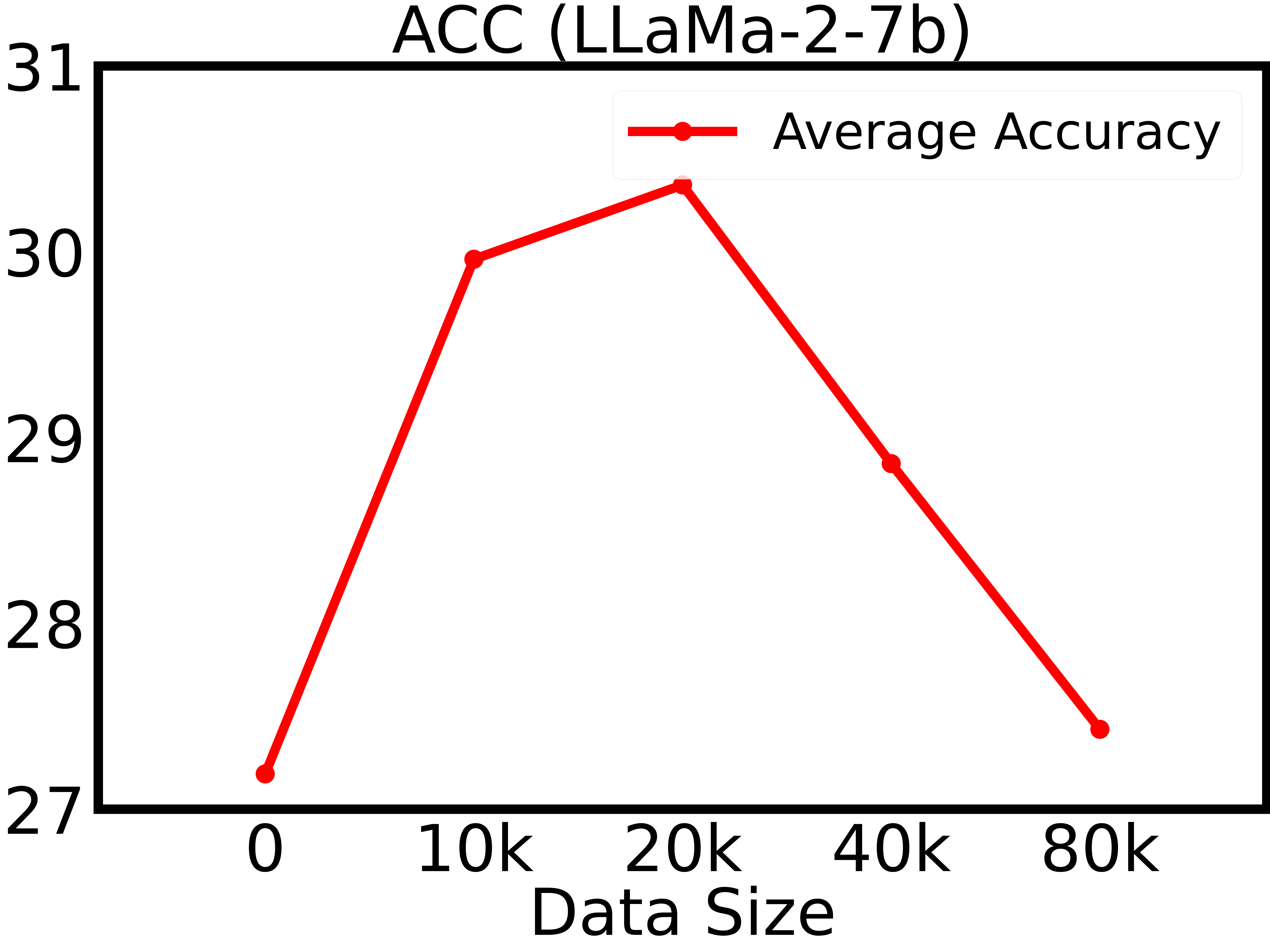}
    \caption{MDIT Scaling(7b)}
\end{subfigure}

\caption{The left figure shows the t-SNE plots of multiple datasets enhanced with MDIT. Red indicates the original data, while blue represents the newly generated data produced by MDIT. The two right figures show the performance scaling of the 1.3B \& 7B model with the MDIT under different $k$ values.}
\label{tab:tsne_scaling}
\end{figure*}

\begin{table*}[ht]
\centering

\begin{tabular}{@{}lcccccc@{}}
\toprule
\textbf{Model} & \multicolumn{1}{c}{\textbf{General QA}} & \multicolumn{1}{c}{\textbf{Math Reasoning}} & \multicolumn{2}{c}{\textbf{Code Generation}} \\ 
& \textbf{ARC Challenge} & \textbf{MMLU-Math} & \textbf{HumanEval}  & \textbf{MBPP} \\
\midrule
MDIT    & \textbf{26.28} & \textbf{29.20}  & \textbf{3.05} & \textbf{4.32}    \\ 
\midrule
w/o (Alpaca $\times$  GSM8K)   & +0.43 & -8.40  & +2.44 & -0.37    \\
w/o (GSM8K $\times$ Codealpaca) & +2.39 & -6.80  & +4.27 & -2.15         \\
w/o (Alpaca $\times$ Codealpaca) & -1.11 & -2.10  & +2.44 & -1.97     \\ 
\bottomrule
\end{tabular}
\caption{Performance of MDIT on the \textsc{Sheared-LLaMA-1.3B} model with different combinations of task interpolation. The main experiment applies pairwise combinations of the Alpaca, GSM8K and CodeAlpaca tasks. The rows below show the performance with the respective combinations removed. "+" indicates performance improvement over the main experiment, while "-" indicates a decline.}
\label{tab:mixup_contribution}
\end{table*}

\subsection{Main experiment}
The main results are shown in Table \ref{tab:main_experiment}. Our \textbf{MDIT} Supervised Fine-Tuning (SFT) model achieved the best average performance among SFT alignment models across different foundation models. For example, in experiments with the \textsc{Sheared-LLaMA-1.3B} model, MDIT improves the average accuracy on four test sets by 2.31\% compared to the original model and outperforms the baseline methods IFD, DEITA and Evol-Instruct by 1.27\%, 1.79\% and 0.97\% respectively. In experiments with the \textsc{LLaMA-2-7B} model, MDIT achieves an even greater improvement of 3.32\% over the original model. Furthermore, on the \textsc{LLaMA-2-13B} model, MDIT improves the average accuracy by 3.9\% compared to the original model.

By generating large amounts of diverse tasks using MDIT and finetuning LLM, we observe performance improvements across tasks such as general question answering, math reasoning, and code generation. The model learns richer semantic representations, showing enhanced generalization capabilities when handling more challenging tasks.

We utilize t-SNE visualization to further illustrate the impact of MDIT on data diversity. As depicted in Figure \ref{tab:tsne_scaling}, the original data primary cluster in the central region of the feature space, while the generated embeddings are more widely dispersed, showing that MDIT creates new and diverse instruction data with richer semantic content. 

\textbf{Data Scaling:} We investigate the impact of data scaling on the \textsc{Sheared-LLaMA-1.3B} and \textsc{LLaMA-2-7B} models by finetuning them with data budgets \emph{m} ranging from 10K to 80K samples. Figure \ref{tab:tsne_scaling} shows that our models outperform the original models across all data scales, with performance gains being most notable when the data volume is relatively small. In particular, the 1.3B model improves as the dataset size increases, while the 7B model initially benefits but eventually declines. It suggests that smaller models require larger datasets for better performance, while larger models perform well with a moderate data scale.

\begin{table*}[htbp]
\centering
\scalebox{0.9}{
\begin{tabular}{cccccc}
\toprule
\textbf{Method} & \textbf{ARC Challenge} & \textbf{MMLU-Math} & \textbf{HumanEval} & \textbf{MBPP} & \textbf{Average} \\
\midrule
$\alpha = 1$ & 28.67 & 22.30 & 5.49 & 5.20 & 15.41 \\
$\alpha = 2$ & 27.47 & 21.40 & 6.71 & 5.55 & 15.28 \\
$\alpha = 4$ & 26.79 & 25.00 & 6.71 & 3.45 & \textbf{15.49} \\
$\alpha = 8$ & 27.73 & 21.80 & 4.88 & 5.53 & 14.99 \\
$\alpha = 12$ & 27.99 & 25.30 & 6.10 & 0.95 & 15.08 \\
\bottomrule
\end{tabular}}
\caption{Performance of MDIT on \textsc{Sheared-Llama-1.3B} model under various $\alpha$ values.}
\label{tab:alpha}
\end{table*}

\begin{table*}[htb]
\centering
\scalebox{0.9}{
\begin{tabular}{cccccc}
\toprule
\textbf{Size} & \textbf{ARC Challenge} & \textbf{MMLU-Math} & \textbf{HumanEval} & \textbf{MBPP} & \textbf{Average} \\
\midrule
10000 & 43.77 & 27.60 & 18.90 & 17.17 & 26.86 \\
+ MDIT & 42.92 & 32.10 & 17.59 & 18.90 & \textbf{27.88} \\
\midrule
20000 & 44.20 & 28.50 & 18.29 & 17.31 & 27.07 \\
+ MDIT & 44.45 & 30.80 & 21.34 & 17.91 & \textbf{28.63} \\
\midrule
40000 & 45.48 & 30.90 & 16.89 & 20.12 & 28.35 \\
+ MDIT & 45.48 & 32.00 & 20.12 & 18.84 & \textbf{29.11} \\
\midrule
80000 & 45.14 & 30.50 & 24.39 & 18.04 & 29.52 \\
+ MDIT & 46.25 & 31.80 & 22.56 & 20.84 & \textbf{30.36} \\
\bottomrule
\end{tabular}}
\caption{Performance of MDIT on \textsc{LLaMA-2-7B} model under various data sizes.}
\label{tab:origin_data_size}
\end{table*}

\subsection{Ablation Study} 
\textbf{Effects of Different Tasks Interpolation:}
To evaluate the impact of task interpolation combinations on MDIT, we selectively remove task pairings. As shown in Table \ref{tab:mixup_contribution}, removing Alpaca \( \times \) GSM8K interpolation improves General QA performance but significantly decreases Math Reasoning, with an 8.40\% drop.  Removing GSM8K \( \times \) CodeAlpaca leads to noticeable improvements in General QA (+2.39\%) and Code Generation (+4.27\%), but harms Math Reasoning (-6.80\%), indicating that this combination is beneficial for tasks requiring complex reasoning. The removal of Alpaca \( \times \) CodeAlpaca causes a slight decline in General QA (-1.11\%) and Math Reasoning (-2.10\%), but boosts HumanEval by +2.44\%, showing its importance for question-answering and reasoning tasks. These results emphasize the importance of carefully selecting dataset pairs for interpolation to achieve a balanced performance across different tasks.

\textbf{Effects of Different interpolation Parameter $\alpha$:}
To explore the impact of the interpolation weight on model performance, we vary the $\alpha$ parameter. The interpolation weight $\lambda$ follows a $\text{Beta}(\alpha, \alpha)$ distribution. As $\alpha$ increases, $\lambda$ becomes more concentrated around 0.5, causing the interpolated samples to move farther from the original samples. We select $\alpha$ values from $\{1, 2, 4, 8, 12\}$, the results are shown in Table \ref{tab:alpha}. Our observations indicate that $\alpha = 4$ achieves the best performance.

\textbf{Effects of Different Data Size:}
To evaluate MDIT in few samples scenarios, we conduct extensive experiments on \textsc{Llama-2-7b}. The experiments utilize a subset of the dataset, with $N = \{10000, 20000, 40000, 80000\}$, where $N = 80000$ represents the full dataset. As shown in Table \ref{tab:origin_data_size}, with only 10K training samples, MDIT improves accuracy by 4.50\% on MMLU-Math and 1.73\% on MBPP, resulting in an average accuracy increase of 1.02\%. With 20K samples, the average accuracy increased by 1.56\%. Notably, training with 10K samples using MDIT outperformed training with 20K samples without augmentation. It shows that even with limited training data, MDIT can effectively generate diverse data, significantly improving LLM performance and maximizing the potential of small-scale datasets.

\begin{figure}[h]
 \centering
\scalebox{1}{
        \includegraphics[height=4cm]{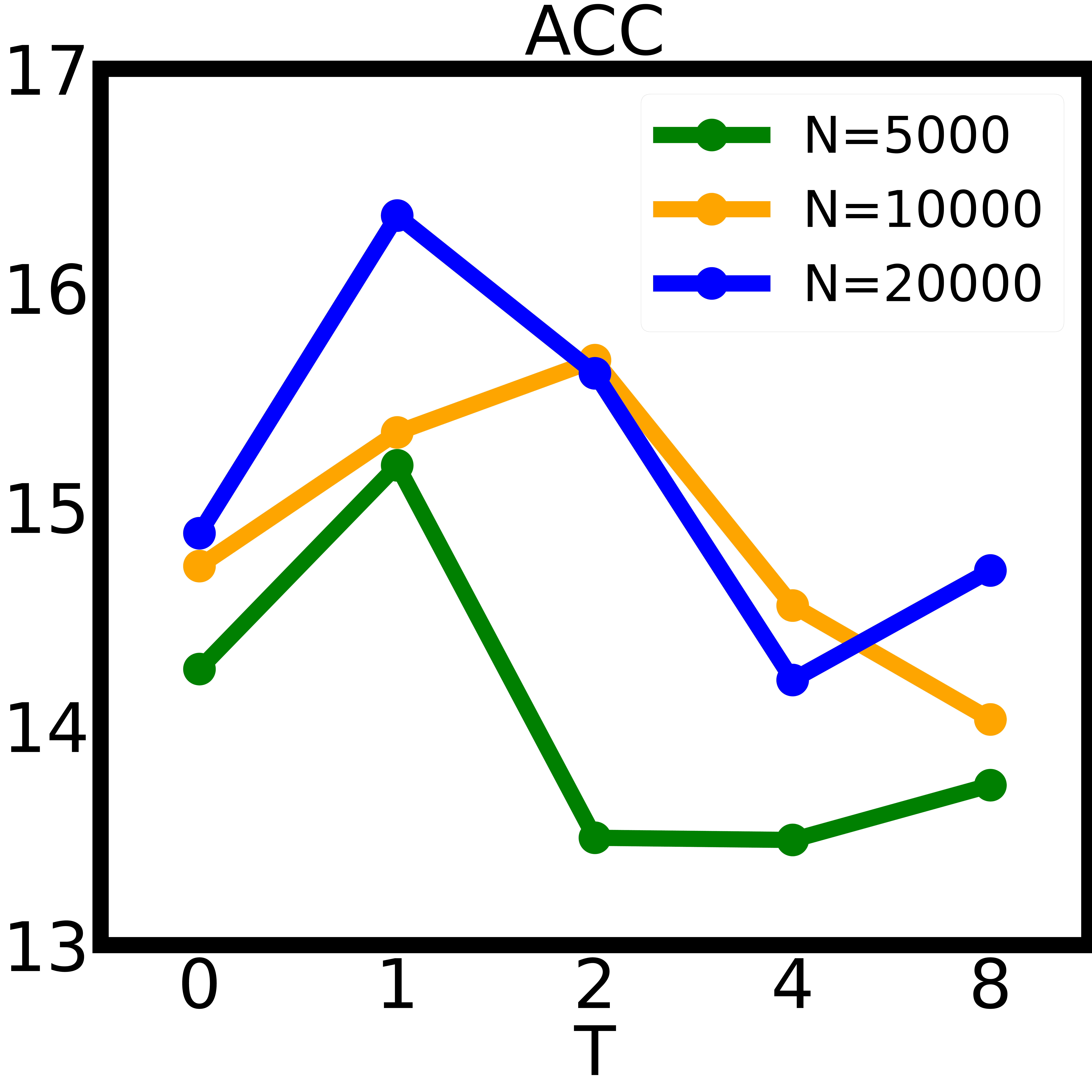}}
        \caption{Performance of MDIT on \textsc{Sheared-Llama-1.3B} model under different generation sample number per original sample pair $T$.}
    \label{T}
\end{figure}

\textbf{Effect of Different Numbers of Generated Samples per Original Sample Pair $T$:}
To evaluate the impact of the number of generated samples per original sample pair $T$ on model performance, we conduct experiments using subsets of the dataset. The values of $T$ are set to $\{0, 1, 2, 4, 8\}$, where $T = 0$ indicates that only the original data is used. As shown in Figure \ref{T}, generating one or two augmented samples per original sample pair leads to improvements in average performance. However, as $T$ increases further, model performance starts to decline, suggesting that there is a limit to the number of useful augmented samples that can be generated from a single original sample pair. Based on these findings, we recommend setting $T$ to no more than 2 for any dataset size.

\section{Conclusion}
In this paper, we propose MDIT, a novel model-free task-level interpolation method that generates diverse tasks for instruction tuning, combined with diversity-based clustering strategies. Extensive experiments show its superior performance that improves the generalization capabilities of LLMs across various tasks. Our method expands the coverage of data within the semantic space, enabling LLMs to learn richer semantic representations. Furthermore, MDIT offers an innovative method to generate diverse instruction data without relying on external resources, providing valuable insights for future research in instruction data management.

\section*{Limitations}
In this paper, we utilize task interpolation to enhance data diversity for instruction tuning. However, even with the application of clustering-based filtering, some noise is inevitably introduced during the data synthesis process. Moving forward, how to implement more effective filtering strategies as well as improve the transparency of the data generation process leaves for future work.

\section*{Ethics Statement}
All the data utilized in our work is gathered from the public resources. We have utilized various open-source models including Sheared-LLaMa-1.3B, LLaMa-2-7B, and LLaMa-2-13B, as well as open-source software such as Hugging Face and PyTorch.

\bibliography{custom}
\end{document}